\documentclass[10pt, a4paper, onecolumn, teaser, showabstract]{styles/naverlabseurope}

\usepackage{lipsum}
\usepackage{multicol}
\usepackage{tikz,pgfplots}
\usepackage{styles/tkz-kiviat}
\usepackage{fontawesome5}
\usepackage{amsmath,graphicx}
\usepackage{amssymb}
\usepackage{bm}
\usepackage{hyperref}
\usepackage{booktabs}
\usepackage{multirow}
\usepackage{xcolor}
\usepackage[htt]{hyphenat}
\usepackage{epstopdf}   % allows EPS with pdflatex (auto-convert)
\usepackage{wrapfig}

\newcommand{\ours}{\textsc{DistilWhisper}}
\newcommand{\ourssp}{\textsc{DistilWhisper}~}

\graphicspath{{figures/}}

\title{\textit{Multilingual DistilWhisper}: Efficient Distillation of Multi-Task Speech Models via Language-Specific Experts}
\titlerunning{\textit{Multilingual DistilWhisper: Efficient Distillation of Multi-Task Speech Models via Language-Specific Experts}}

\correspondingauthor{[thomas.palmeira,marcely.zanon-boito]@naverlabs.com}
\authors{Thomas Palmeira Ferraz$^{\diamondsuit\clubsuit}$ $\quad\quad$ Marcely Zanon Boito$^{\diamondsuit}$ $\quad\quad$ Caroline Brun$^{\diamondsuit}$ $\quad\quad$ Vassilina Nikoulina$^{\diamondsuit}$ \vspace{10pt}}

\affiliations{$^\diamondsuit$ NAVER LABS Europe, France\\$^\clubsuit$ Télécom Paris, Institut Polytechnique de Paris, France}
\website{\faGithub ~ github.com/naver/multilingual-distilwhisper}
\websiteref{\href{https://github.com/naver/multilingual-distilwhisper}}

\teaserfig{\vspace{5pt} \includegraphics[width=0.90\linewidth]{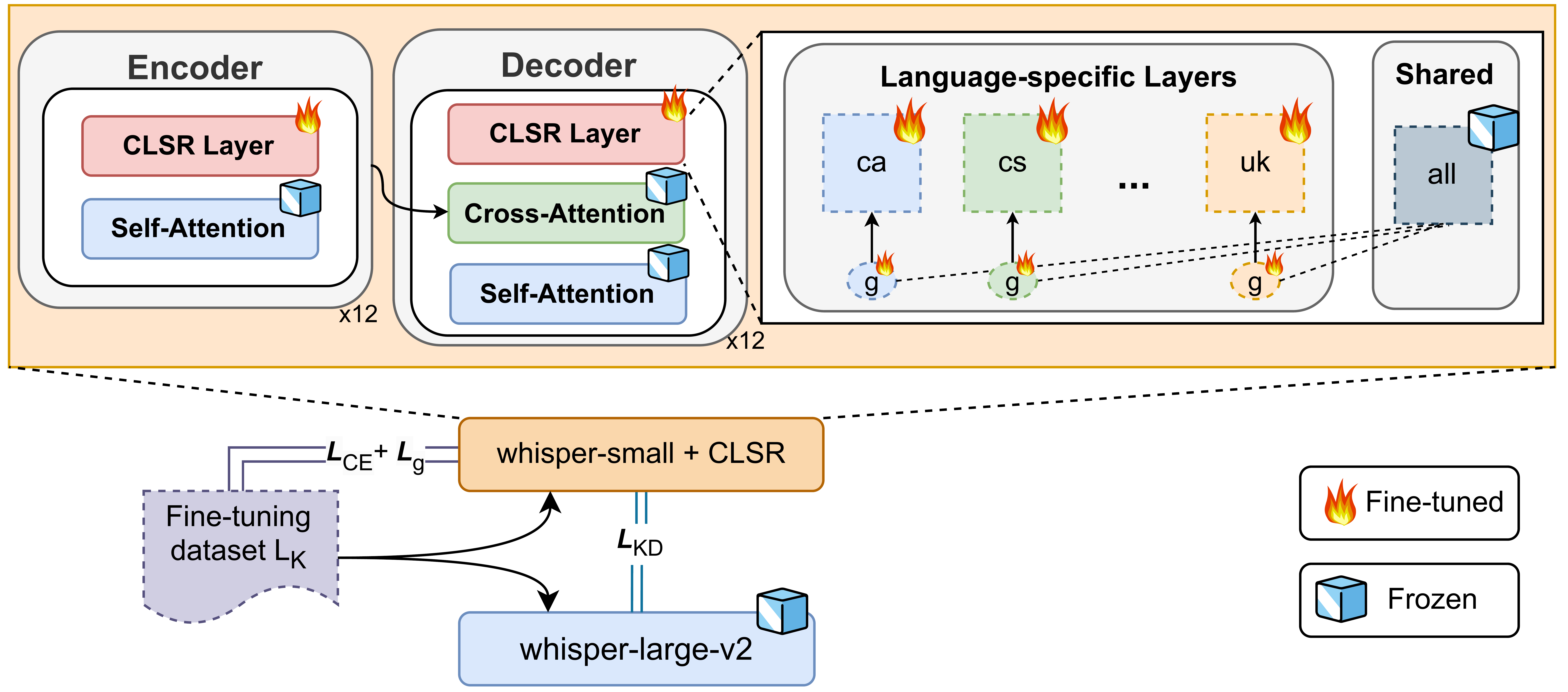} \vspace{2pt} }

\teasercaption{The Multilingual~\textbf{\ours} architecture~(top) and optimization framework~(bottom). \textbf{Architecture:} We extend \texttt{whisper-small} by replacing its feed-forward network (FFN) modules with Conditional Language-Specific Routing~(CLSR) modules in both encoder and decoder. Each CLSR module contains language-specific gates~($g$) that route tokens through either frozen multilingual representations initialized with previous FFN block~(\textit{shared}) or learnable language-specific modules~(\textit{LS}). \textbf{Training:} Our dual optimization combines ASR supervised fine-tuning, gate budget and knowledge distillation from frozen robust teacher~(\texttt{whisper-large-v2}). \label{fig:arch} \vspace{5pt}}

\begin{abstract}

\textit{Whisper} is a multitask and multilingual speech model covering 99 languages. 
It yields commendable automatic speech recognition~(ASR) results in a subset of its covered languages, but the model still underperforms on a non-negligible number of under-represented languages, a problem exacerbated in smaller model versions.
In this work, we propose \ours, an approach able to bridge the performance gap in ASR for these languages while retaining the advantages of multitask and multilingual capabilities.
Our approach involves two key strategies: lightweight modular ASR fine-tuning of \texttt{whisper-small} using language-specific experts, and knowledge distillation from \texttt{whisper-large-v2}. This dual approach allows us to effectively boost ASR performance while keeping the robustness inherited from the multitask and multilingual pre-training. 
Results demonstrate that our approach is more effective than standard fine-tuning or LoRA adapters, boosting performance in the targeted languages for both in- and out-of-domain test sets, while introducing only a negligible parameter overhead at inference.
\vspace{-20pt}
\end{abstract}

\begin{document}

\maketitle

\onecolumn
\section{Introduction}
\label{sec:intro}

\textit{Whisper}~\citep{radford2023robust} is a popular multilingual and multitask speech model that is known for its robustness~(i.e. invariant performance over different out-of-domain data) for automatic speech recognition~(ASR)~\citep{gandhi2022esb}. 
This model covers 99 languages, and jointly trains on ASR, speech translation~(many-to-English), language identification, and voice activity detection tasks. The original paper attributes this multitask training as a reason for the observed robustness of the model to out-of-domain data: compared to the English wav2vec~2.0 model~\citep{baevski2020wav2vec}, Whisper performance seems to generalize better to unseen domains.
Available in many sizes~(from tiny to large-v2), Whisper exhibits an important gap in ASR performance between \texttt{whisper-large-v2}~(largest model) and \texttt{whisper-small}~(second smallest model) on a large set of languages, including low-resource languages, but also many high- and mid-resource ones. This phenomenon in NLP is often referred as \textit{curse of multilinguality}, where the performance drop due to the growing amount of covered languages can only be recovered via extensive model scaling~\citep{mmmt_google,conneau2020unsupervised, goyal2021largerscale}. Such scaling comes with an important inference cost increase: for instance, \texttt{whisper-large-v2} is 2-3 times slower than \texttt{whisper-small}.

A common approach to efficient inference is distilling knowledge from a large multilingual teacher model into a smaller model \citep{sanh2020distilbert,mohammadshahi-etal-2022-small}. However, applying knowledge distillation~(KD) to \texttt{whisper-large-v2}, the best and largest Whisper model, presents a challenge. We would ideally need access to training data across all tasks and languages to preserve robustness, but such data is unavailable, making it challenging to maintain the model's out-of-domain generalization capabilities.

In another direction, \citet{pfeiffer-etal-2022-lifting} and \citet{pratap2023scaling} have demonstrated that the \textit{curse of multilinguality} can also be solved by equipping a moderately sized model with language-specific~(LS) modules. Such architectures allow extending model parameters via extra modules when more languages are added into the model, thus maintaining consistent performance across languages, with no (or very low) extra computations at inference. 

Inspired by these findings, we propose \ours, which extends \texttt{whisper-small} with language-specific feed-forward modules, that are used in parallel with the original feed-forward layers of the model. In order to  preserve the robustness of the original model, \ourssp introduces the following extensions of previous works:
\begin{enumerate}
\item Following \citet{zhang2021share}, we extend conditional language-specific routing~(CLSR) modules with the gating mechanism that can route input representation either through the original feed-forward module or through newly learned LS feed-forward module;
\item When learning language-specific modules, we use \texttt{whisper-large-v2} as a teacher model with the hypothesis that the KD loss should help reproduce the robustness of the larger Whisper model.
\end{enumerate}

Through extensive experiments on a diverse set of languages we demonstrate the effectiveness of \ourssp compared to standard fine-tuning or LoRA adapters~\citep{hu2021lora}. Our lightweight ASR fine-tuning approach based on CLSR modules generalizes better than LoRA, and the introduction of KD further boosts results in both in- and out-of-domain test sets.
We perform additional ablation studies showing our approach can cope with different amounts of training data. Finally, we demonstrate that the flexibility introduced by the gating mechanism equips \ourssp with an efficient adaptation approach, leveraging the language-specific modules only when those are relevant. 
We make available the models' weights\footnote{Weights available at: \url{https://huggingface.co/collections/naver/multilingual-distilwhisper-6576ecae8d209fc6a767d9e7}.}  and code\footnote{Code available at: \url{https://github.com/naver/multilingual-distilwhisper}.} 
developed in this work.

\section{Background}
\label{sec:background}

\subsection{State of the art for ASR}
Current approaches for ASR mainly rely on the adaptation of pre-trained Transformer stacks learned through self-supervision~(i.e. SSL models) on unlabeled audio data. Such pre-trained models vary on the usage of pretext tasks~\citep{baevski2020wav2vec,hsu2021hubert,chen2022wavlm} and language coverage~\citep{conneau21_interspeech,babu22_interspeech, pratap2023scaling,zhang2023google}. 
In contrast to this branch of research, the Whisper model relies on weak supervision, which means that the architecture is trained on weakly labeled data only~(no self-supervision). Nonetheless, \citet{radford2023robust} show that with sufficient amounts of data, the Weakly Supervised Whisper model reaches competitive results compared to monolingual and multilingual SSL models~\citep{gandhi2022esb,pratap2023scaling}.

\subsection{Knowledge distillation}

\setlength\parindent{0pt}\textbf{Knowledge distillation~(KD)} has been initially proposed by~\citet{kd_hinton} to distill knowledge from ensemble of models into a single model for ASR. It has further been used to distill knowledge from a large teacher model into smaller student models~\citep{sanh2020distilbert,mohammadshahi-etal-2022-small, shen2023language}. While original KD methods relied on minimization of KL-divergence between a teacher model and a student model, \citet{js_vs_kd_acl23} and \citet{js_vs_kd_bb2023} have recently shown that symmetric divergences, such as Jensen-Shannon~(JS) divergence, suffer less from borderline behaviors and lead to better results on sequence level distillation. 

\subsection{Parameter-efficient Fine-tuning}
\setlength\parindent{0pt}\textbf{Adapters} are small lightweight modules which are commonly used in NLP to adapt pre-trained models to new tasks or domains. In speech-related tasks, adapter-based fine-tuning has been utilized for speech translation~\citep{le2021lightweight,gow2023naver,antonios2022findings}, and domain adaptation~\citep{thomas2022efficient,tomanek2021residual}, for which they exhibit a similar performance to standard fine-tuning, but with only a fraction of trainable parameters. 
We also find work on task-adaptation of Whisper~\citep{radhakrishnan2023parameter,wang23ga_interspeech,feng2023peft} using LoRA adapters. In contrast to adapters, in this work we introduce gated LS modules into Whisper, and propose a parameter-efficient KD approach that allows us to increase robustness to out-of-domain data.

\section{DistilWhisper}
\label{sec:CLSR}

With the goal of increasing performance for different languages in models of limited capacity, we propose the \ourssp approach: we plug conditional language-specific routing~(CLSR) modules~\citep{zhang2021share} into a small Whisper~(\texttt{whisper-small}), and optimize these modules jointly on ASR fine-tuning and KD from a larger Whisper~(\texttt{whisper-large-v2}). Figure~\ref{fig:arch} presents our architecture, below we detail its key components.

\subsection{Conditional Language-Specific Routing~(CLSR) module}
\label{subsec:clsr}

We extend CLSR modules for the first time to the speech domain. This module learns a hard binary gate g(·) for each input token by using its hidden embedding $z^l$. These decisions enable a layer to selectively guide information through either a Language-Specific path denoted as $h^{lang}$ or a shared path referred to as $h^{shared}$, as in Eq~\ref{eq:clsr}:

\begin{equation}\label{eq:clsr}
    \text{CLSR}(z^l) = g(z^l)\cdot h^{lang}(z^l) + (1 - g(z^l))\cdot h^{shared}(z^l). 
\end{equation}

In contrast to the original CLSR, in this work we use language-specific gates as shown in Figure~\ref{fig:arch}, instead of sharing them across languages. This allows us to train LS modules individually~(i.e. in parallel), and then only load the relevant modules at inference. Moreover, our approach also differs from the original CLSR by the positioning: supported by \citet{zhang2021share} and \citet{pfeiffer-etal-2022-lifting} works, we limit CLSR to the feed-forward, which we entirely replace with the CLSR module, reducing further the number of parameters. 

Following~\citet{zhang2021share}, the gating mechanism is implemented as follows: each gate $g(.)$ consists of a two-layer bottleneck network, to which we add an increasing zero-mean Gaussian noise during training to facilitate discretization and enable gradient flow through the binary decisions. At inference time, we adopt hard gating, where the gate outputs are deterministically set to either 0 or 1 based on the learned routing decisions.

\subsection{\ourssp approach}
\label{subsec:approach}
\textbf{\ours} approach is detailed in Figure~\ref{fig:arch}. Our student is enriched with CLSR modules at each feed-forward for each language. These CLSR layers are initialized from the frozen weights of the corresponding feed-forward layer. At training time, for each language the model updates only the corresponding language-specific modules and gates. At inference time, the model loads the shared modules~(multilingual) and the LS modules and gates for the languages of interest, resulting in a limited parameter overhead. We highlight that the use of CLSR modules brings more flexibility to our architecture when compared to adapters, as it allows for routing at the token-level. This makes this approach more capable of leveraging pre-existing knowledge~(shared frozen module) via LS gating activation.

\subsection{\ourssp optimization}
\label{subsec:optimization}
Following~\citet{zhang2021share}, when learning CLSR module parameters, in addition to standard cross-entropy loss $\mathcal{L}_{\text{CE}}$, we employ a gate budget loss $\mathcal{L}_{\text{g}}$ (Eq~\ref{eq:gateloss}) to balance models' usage of LS and language-shared modules. 

The gate budget loss relies on the gate $g(.)$ activation values for a pair (audio, text) $(X,Y)$ in a batch $\mathcal{B}$, which is expressed by:
\begin{equation}
\mathcal{G}_{(X,Y)} = \sum_{x \in X}\sum_{m \in \mathcal{M}_{\text{enc}}} g_m (x) + \sum_{y \in Y}\sum_{m \in \mathcal{M}_{\text{dec}}} g_m (y)
\end{equation}
where $\mathcal{M}_{\text{enc}}$ and $\mathcal{M}_{\text{dec}}$ are respectively the encoders and decoders layers, and $g_m(.)=1$ when LS module is selected, or $0$ otherwise. The average of this gate usage is constrained to a budget $b$~(Eq~\ref{eq:gateloss}):

\begin{equation}\label{eq:gateloss}
    \mathcal{L}_{\text{g}} = \left\lvert \frac{\sum_{(X,Y) \in \mathcal{B}} \mathcal{G}_{(X,Y)}}{\sum_{(X,Y) \in \mathcal{B}} (|X||\mathcal{M}_{\text{enc}}| + |Y||\mathcal{M}_{\text{dec}}|)} - b \right\rvert,
\end{equation}

For KD, following \citet{js_vs_kd_acl23} and \citet{js_vs_kd_bb2023}, we use JS divergence, whose loss is detailed in Eq~\ref{eq:kd}: 

\begin{equation}\label{eq:kd}
    \mathcal{L}_{\text{KD}} = \frac12 \mathbb{E}_{\bm{Y} \sim p} \left[ \log \tfrac{p(\bm{Y})}{m(\bm{Y})} \right] + \frac12 \mathbb{E}_{\bm{Y}' \sim q_\theta} \left[ \log \tfrac{q_\theta (\bm{Y}')}{m(\bm{Y}')} \right]
\end{equation}

where $p$ is the teacher distribution, $q_\theta$ is the student distribution, $\bm{Y}$ and $\bm{Y}'$ are sampled from the teacher's and student's distributions and compared with their average $m(\cdot)=\frac12 p(\cdot) + \frac12q_\theta(\cdot)$. 

Thus, CLSR modules parameters are learned to minimize final loss expressed as:

\begin{equation}\label{eq:distilwhisperloss}
    \mathcal{L} = \mathcal{L}_{\text{CE}} + \mathcal{L}_{\text{g}} + \alpha\mathcal{L}_{\text{KD}}.
\end{equation}
\section{Experimental Setup}\label{sec:setup}

\subsection{Datasets}
We downsample the train and validation sets of the CommonVoice~13.0~(CV-13) dataset~\citep{ardila-etal-2020-common}, using equal amounts of training data for each selected language: 10k utterances for training~(approx. 14\,h), 1k for validation. Data selection depends on the amount of up-votes utterances received by annotators. We do not downsample the test set.
The FLEURS~\citep{conneau2023fleurs} dataset is used for out-of-domain evaluation, as it provides both a good language overlap with CV-13, and an effective out-of-domain setting for ASR evaluation. For instance, average number of tokens per sample for CV-13 is 36, and 97 for FLEURS.

\subsection{Language Selection}
We consider all Whisper languages with a WER gap of more than 11 between large and small models on CV-13.
We then narrow this list considering: 1) minimum amount of utterances~(10k); 2) overlap with the FLEURS dataset. The final list of languages is: Catalan~(ca), Czech~(cs), Galician~(gl), Hungarian~(hu), Polish~(pl), Thai~(th), Tamil~(ta) and Ukrainian~(uk).\footnote{Although Arabic would also qualify considering our criteria, we find that the dialect from FLEURS differs from the ones present on CV-13.} These languages encompass 5 language sub-families and vary widely in terms of coverage in the available Whisper training data, spanning from 4,300\,h~(pl) to just 9\,h~(gl).

\subsection{Models}
We compare our approach to both \texttt{whisper-small}~(pre-trained student) and \texttt{whisper-large-v2} (teacher) models, as well as two approaches of fine-tuning~(FT) for the student: standard fine-tuning~(all weights are updated), and LoRA adaptation on top of the feed-forward layer. Finally, we also investigate the impact of the CLSR layer without the use of KD~(CLSR-FT), decoupling the effect of KD from the flexibility offered by the routing mechanism on the consequent robustness of the model.

\subsection{Implementation details}
We train all models using the Transformers library~\citep{wolf2020transformers}, using the pre-trained weights for \texttt{whisper-small} and \texttt{whisper-large-v2} available on HuggingFace\footnote{Models weights are available at: \url{https://huggingface.co/openai/whisper-small} and \url{https://huggingface.co/openai/whisper-large-v2}}. All models are trained for $10$ epochs using a learning rate of $10^{-4}$ with linear decay, one epoch of warm-up, batch size of 16, and label smoothing factor of 0.1. For LoRA, we use the hyperparameters proposed by~\citet{wang23ga_interspeech}. For CLSR training, we set the gate budget $b = 0.5$ and skip-gate probability $s = 0.2$. For knowledge distillation (KD), we employ the JS divergence with temperature $\tau = 1$. The full learning objective for our experiments is given by:

\begin{equation}
    \mathcal{L} = \mathcal{L}_{\text{CE}} + \mathcal{L}_{\text{g}} + 2\mathcal{L}_{\text{KD}}
\end{equation}

We report normalized WER using the Whisper normalization procedure, with a slight modification to avoid splitting numbers and latin-scripted text into individual characters for languages that do not use space delimitation (e.g., Thai).
In all cases, the best model is chosen based on WER on the down-sampled CV-13 validation set. 

\begin{table}
    \centering
    \resizebox{\textwidth}{!}{
    \begin{tabular}{lcc|cccccccc}
    \toprule
    \multicolumn{11}{c}{\textbf{FLEURS (out-of-domain)}} \\\midrule
    \multicolumn{1}{c}{\textbf{Model}} & \textbf{\# params} & \textbf{avg} & \textbf{ca} & \textbf{cs} & \textbf{gl} & \textbf{hu} & \textbf{pl} & \textbf{ta} & \textbf{th} & \textbf{uk} \\\midrule
    \texttt{whisper large-v2} & 1.5 B & \multicolumn{1}{c|}{12.5} & 5.6 & 14.3 & 16.6 & 17.9 & 5.9 & 19.3 & 12.2 & 8.1 \\
    \texttt{whisper-small} & 244 M & \multicolumn{1}{c|}{28.3} & \textbf{14.6} & 40.4 & 32.7 & 43.0 & \textbf{16.7} & 36.0 & 22.8 & \textbf{20.5} \\ \midrule
    \texttt{whisper-small} + FT & 244 M & \multicolumn{1}{c|}{23.3\scriptsize{$\pm$0.06}} & 15.5 & 31.0 & 16.9 & \textbf{36.7} & 22.0 & 22.7 & 15.6 & 25.9 \\
    \texttt{whisper-small} + LoRA-FT & 379 M & \multicolumn{1}{c|}{24.9\scriptsize{$\pm$0.07}} & 17.6 & 36.9 & 18.2 & 41.6 & 25.9 & \textbf{15.2} & \textbf{11.7} & 31.8 \\\midrule
    \texttt{whisper-small} + CLSR-FT & 369 M & \multicolumn{1}{c|}{23.4\scriptsize{$\pm$0.19}} & 15.7 & 30.5 & 17.2 & 36.9 & 22.8 & 22.7 & 15.6 & 25.8 \\
    \ours& 369 M & \multicolumn{1}{c|}{\textbf{22.8}\scriptsize{$\pm$0.21}} & \textbf{15.3} & \textbf{30.2} & \textbf{16.7} & 36.9 & \textbf{21.4} & 21.8 & 15.1 & \textbf{24.9} \\\bottomrule
    \\\toprule
    \multicolumn{11}{c}{\textbf{Common Voice 13.0 (in-domain for FT only)}} \\\midrule
    \multicolumn{1}{c}{\textbf{Model}} & \textbf{\# params} & \textbf{avg} & \textbf{ca} & \textbf{cs} & \textbf{gl} & \textbf{hu} & \textbf{pl} & \textbf{ta} & \textbf{th} & \textbf{uk} \\\midrule
    \texttt{whisper large-v2} & 1.5 B & \multicolumn{1}{c|}{{\color[HTML]{9B9B9B} 14.9}} & {\color[HTML]{9B9B9B} 16.9} & {\color[HTML]{9B9B9B} 14.4} & {\color[HTML]{9B9B9B} 18.9} & {\color[HTML]{9B9B9B} 18.7} & {\color[HTML]{9B9B9B} 8.0} & {\color[HTML]{9B9B9B} 17.3} & {\color[HTML]{9B9B9B} 9.2} & {\color[HTML]{9B9B9B} 15.5} \\
    \texttt{whisper-small} & 244 M & \multicolumn{1}{c|}{{\color[HTML]{9B9B9B} 31.4}} & {\color[HTML]{9B9B9B} 30.1} & {\color[HTML]{9B9B9B} 38.4} & {\color[HTML]{9B9B9B} 35.5} & {\color[HTML]{9B9B9B} 45.6} & {\color[HTML]{9B9B9B} 18.6} & {\color[HTML]{9B9B9B} 30.0} & {\color[HTML]{9B9B9B} 20.3} & {\color[HTML]{9B9B9B} 32.3} \\ \midrule
    \texttt{whisper-small} + FT & 244 M & \multicolumn{1}{c|}{16.3\scriptsize{$\pm$0.09}} & \textbf{13.7} & 20.5 & \textbf{11.3} & 24.1 & 16.3 & 13.6 & 7.4 & 23.4 \\
    \texttt{whisper-small} + LoRA-FT & 379 M & \multicolumn{1}{c|}{18.2\scriptsize{$\pm$0.02}} & 14.0 & 23.7 & 12.7 & 28.0 & 21.2 & \textbf{12.0} & 7.9 & 26.4 \\\midrule
    \texttt{whisper-small} + CLSR-FT & 369 M & \multicolumn{1}{c|}{16.3\scriptsize{$\pm$0.08}} & 14.1 & 20.3 & 11.6 & 24.3 & 16.1 & 13.3 & 7.4 & 23.4 \\
    \ours& 369 M & \multicolumn{1}{c|}{\textbf{16.0}\scriptsize{$\pm$0.04}} & \textbf{13.8} & \textbf{20.0} & 11.8 & \textbf{24.0} & \textbf{15.9} & 12.6 & \textbf{7.2} & \textbf{23.1} \\\bottomrule
    \end{tabular}}
    % \caption{WER~($\downarrow$) with dataset averages~(avg) for baselines~(top), adaptation approaches~(middle), and our method~(bottom) for in-domain~(CV-13, FT only) and out-of-domain~(FLEURS, all) test sets. Best results for \texttt{whisper-small} in \textbf{bold}.}

    \caption{WER~($\downarrow$) for both in-domain and out-of-domain evaluation settings. 
    \textbf{Top panel:} \textbf{FLEURS} (out-of-domain).
    \textbf{Bottom panel:} \textbf{Common Voice 13.0 (CV-13)} (in-domain only for fine-tuned~(FT) models) - Results of pre-trained models on CV-13 are shown in {\color[HTML]{9B9B9B}\textbf{gray}}, as they are not directly comparable to FT models. 
    Each panel reports the number of active parameters, dataset average WER~(mean{\footnotesize{ $\pm$std}}) and per-language WER. Rows are grouped as \emph{baselines}~(top), \emph{adaptation approaches}~(middle), and \emph{our method}~(bottom), with FT-only~(CLSR-FT) or with distillation~(\ours).
    Best results for fine-tuned \texttt{whisper-small} as well as cases where pre-FT models performed better are shown in \textbf{bold}.}
    \label{tab:results1}
\end{table}
\begin{table}[]
\centering
\begin{tabular}{lccc|ccc|ccc}\toprule
                      & \textbf{Train} & \textbf{FLEURS}    & \textbf{CV-13}     & \multicolumn{3}{c}{\textbf{FLEURS}} & \multicolumn{3}{c}{\textbf{CV-13}} \\
                      & \textbf{size}  & \textbf{avg}       & \textbf{avg}       & \textbf{ca}      & \textbf{ta}      & \textbf{th}     & \textbf{ca}      & \textbf{ta}      & \textbf{th}    \\\midrule
\texttt{whisper-small}+CLSR-FT & 3k    & 20.5\scriptsize{$\pm$0.17} & 15.0\scriptsize{$\pm$0.07} & 17.9    & 25.6    & 18.0   & 19.0    & 16.4    & 9.8   \\
\ours         & 3k    & \textbf{20.2}\scriptsize{$\pm$0.13} & \textbf{14.6}\scriptsize{$\pm$0.08} & \textbf{17.4}    & 25.5    & \textbf{17.7}   & \textbf{18.7}    & \textbf{15.7}    & \textbf{9.6}   \\\midrule
\texttt{whisper-small}+CLSR-FT & 10k   & 18.0\scriptsize{$\pm$0.25} & 11.6\scriptsize{$\pm$0.01} & 15.7    & 22.7    & 15.6   & 14.1    & 13.3    & 7.4   \\
\ours         & 10k   & \textbf{17.4}\scriptsize{$\pm$0.13} & \textbf{11.2}\scriptsize{$\pm$0.08} & \textbf{15.3}    & \textbf{21.8}    & \textbf{15.1}   & \textbf{13.8}    & \textbf{12.6}    & \textbf{7.2}   \\\midrule
\texttt{whisper-small}+CLSR-FT & 28k   & 15.7\scriptsize{$\pm$0.15} & 9.5\scriptsize{$\pm$0.13}  & 13.5    & 19.8    & 13.9   & 11.3   & 11.3    & 6.0   \\
\ours         & 28k   & \textbf{15.5}\scriptsize{$\pm$0.03} & \textbf{9.3}\scriptsize{$\pm$0.06}  & \textbf{13.3}    & \textbf{19.3}    & \textbf{13.7}   & 11.3    & \textbf{11.0}    & \textbf{5.7}  \\\bottomrule
\end{tabular}
\caption{Average WER~($\downarrow$) for models trained with different training data sizes~(3k, 10k, and 28k utterances). Results are reported on in-domain~(\textbf{CV-13}) and out-of-domain~(\textbf{FLEURS}) test sets, including per-language scores for the languages with at least 28k utterances at training split (\texttt{ca}, \texttt{ta}, and \texttt{th}). Best results for each training size are shown in \textbf{bold}.}
\label{tab:results2}
\end{table}

\section{Results}
\label{sec:results}

We conduct training for each setting using three distinct seeds and present the average scores.
Table~\ref{tab:results1} presents our results. The top portion presents 
\texttt{whisper-large-v2} (upper bound) and \texttt{whisper-small} (lower bound) pre-trained scores. The middle portion presents standard fine-tuning~(FT) and LoRA adaptation at the feed-forward layers~(LoRA-FT). Our results are presented in the bottom: CLSR-FT corresponds to the setting without $\mathcal{L}_{\text{KD}}$, while \ourssp is the complete setting in which both CLSR and KD losses are leveraged.

\subsection{\ourssp vs. other adaptation approaches} 
For \texttt{whisper-small}, we observe that both FT and LoRA-FT approaches~(middle portion of Table~\ref{tab:results1}) are able to improve performance on both in- and out-of-domain test sets. 
However, FT achieves this improvement at the cost of language specialization, reducing performance in other languages. In contrast to that, LoRA-FT is a light adaptation technique that does not modify the pre-trained representation. This method increases performance on both in-domain~(avg -13.1) and out-of-domain~(avg~-3.5) test sets compared to \texttt{whisper-small}. \ourssp further improves performance over \texttt{whisper-small}~(avg -15.3) and LoRA-FT~(avg -2.2) for in-domain data. It also presents better out-of-domain adaptation capabilities compared to LoRA-FT~(avg~-2.1).

\subsection{Impact of knowledge distillation}
We observe that \ourssp on average outperforms all other adaptation approaches~(FT, LoRA-FT) for in- and out-of-domain test sets~(bottom portion of Table~\ref{tab:results1}). 
Comparing our models~(CLSR-FT and \ours), we
observe that the version with KD~(\ours) exhibits a slight increase in average in-domain performance~(-0.3). In out-of-domain settings, this model consistently outperforms CLSR-FT across all languages~(avg -0.6), which confirms our initial hypothesis that the KD loss leverages the robustness from the teacher into the final model. 
Overall, these results highlight the effectiveness of our proposed architecture: we reduce the out-of-domain performance gap between \texttt{whisper-large-v2} and \texttt{whisper-small} by $35.2\%$~(avg~-5.5) with a parameter overhead at inference time of only $10\%$~(25~M).

\subsection{Effect of training data size} 
We now show the effectiveness of our approach on lower and higher data resource settings.
For this, we select a subset of languages for which we find more training data available on CV-13~(ca, th, ta). Table~\ref{tab:results2} presents results for our approach in low~(3k utterances; $\sim$4\,h), and higher-resource settings~(28k utterances; $\sim$40\,h), compared to the 10k results from Table~\ref{tab:results1}.
We observe that, as expected, increasing the amount of trainable examples leads to superior ASR performance for both approaches, with the leveraging of KD~(\ours) being consistently superior to CLSR-FT. For the 28k setup~(ca, th, ta), we reduce the out-of-domain WER gap between \texttt{whisper-large-v2} and \texttt{whisper-small} by~$75\%$~(from $12$ to $3$~WER).\footnote{\texttt{whisper-large-v2} and \texttt{whisper-small} avg FLEURS scores for ca, th, ta are respectively 12.5 and 24.5.} 
For the 3k setup, we reduce the WER gap by 35.8\% using only 4\,h of training data.
This implies that our approach has the potential to improve ASR performance across low-resource languages for which less training data is available.

\begin{figure}
    \centering
    \includegraphics[width=0.9\linewidth]{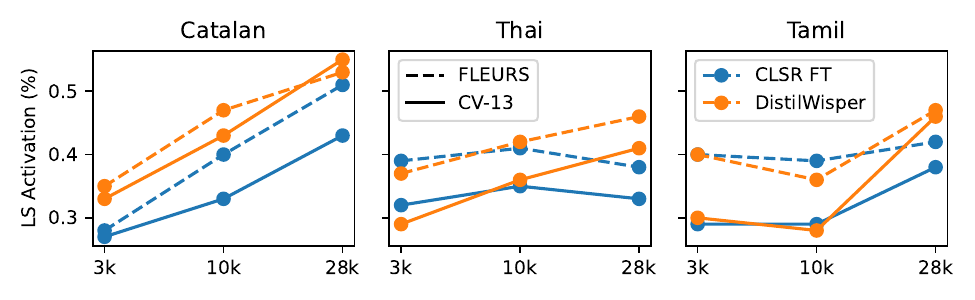}
    \caption{Ratio of language-specific~(LS) expert activations selected by each model across different training data sizes~(x-axis), evaluation domains — in-domain~(\textbf{CV-13}, solid lines) and out-of-domain~(\textbf{FLEURS}, dashed lines) — and languages. We compare the direct baseline~(\textcolor[HTML]{1f77b4}{\textbf{CLSR-FT}}) and with our knowledge-distilled CLSR-FT+KD~(\textcolor[HTML]{ff7f0e}{\textbf{\ours}}).}
    \label{fig:gates}
\end{figure}

\subsection{Gate Activation Analysis}
To better understand how the model uses routing mechanism, we plot gate activation statistics for both CLSR-FT and \ourssp in Figure~\ref{fig:gates}. We observe that the models tend to rely more on the new language-specific modules in out-of-domain settings~(FLEURS vs CV-13), which could be attributed to the greater complexity and larger size of sentences in FLEURS.
Also, as expected, increasing the training data size leads to more reliable LS modules, and therefore higher LS usage. The only exception for this is Thai at the 28k setup, and this might be due to dataset quality and requires further investigation. When comparing the 3 languages, we observe that Catalan exhibits a higher reliance on LS routes, which could also be related to the data quality for this language in CV-13. Finally, we observe that for languages with a weaker teacher~(Thai, Tamil) the model may receive contradictory signals at lower-resource settings~(3k, 10k), leading to less LS routing usage with KD. However, in the higher resource setting~(28k), KD usage leads systematically to more reliable LS module and therefore higher LS routing.
\section{Conclusion}
\label{sec:conclusion}

We presented \ours, a parameter-efficient distillation approach that boosts performance of \texttt{whisper-small} by leveraging the robustness from the \texttt{whisper-large-v2} into a smaller model, while preserving its multilingual capabilities. This is done by adding language-specific gated modules, and by jointly optimizing ASR fine-tuning and KD losses. Compared to LoRA adapters, and across eight languages, we are able to consistently improve performance in both in- and out-of-domain test sets, while adding only a negligible number of parameters at inference time. We believe this architecture makes Whisper models more accessible to researchers and practitioners, as it boosts the performance of a low-inference cost model by 35.2\% using only 14\,h of training data.

\section*{Acknowledgements}

This research was partially supported by the French \textit{Agence Nationale de la Recherche, ANR} as part of the project \textbf{Diké - Bias, Fairness and Ethics of Compressed Language Models}, under grant number \texttt{ANR-21-CE23-0026-02}. % A CC-BY public copyright license has been applied by the authors to the present document and will be applied to all subsequent versions up to the Author Accepted Manuscript arising from this submission, in accordance with the grant's open access conditions.

\vspace{0.5em}
\noindent
\begin{minipage}[t]{0.15\textwidth}
  \vspace{0pt}
  \includegraphics[width=\linewidth,keepaspectratio]{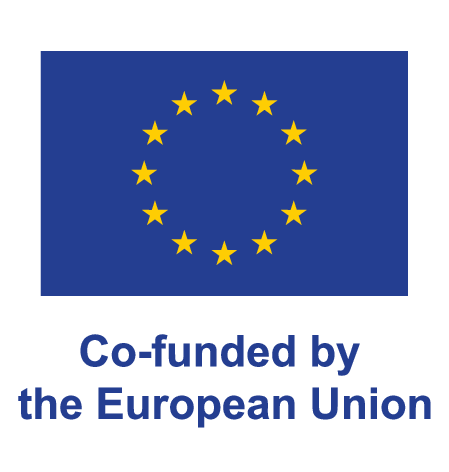}
\end{minipage}\hfill
\begin{minipage}[t]{0.80\textwidth}
  \vspace{0pt}
  This work was also co-funded by the European Union's Horizon Europe Research and Innovation programme through the project \textbf{UTTER – Unified Transcription and Translation for Extended Reality}\footnote{For more information please visit: \url{https://he-utter.eu/}.} under Grant Agreement No. \texttt{101070631}.

  \medskip\footnotesize
  Funded by the European Union. Views and opinions expressed are however those of the author(s) only and do not necessarily reflect those of the European Union. Neither the European Union nor the granting authority can be held responsible for them.
\end{minipage}

\vspace{1em}

\bibliographystyle{styles/plainnat}
\bibliography{references}

\appendix
\vspace{20pt}
\section{Effect of Temperature and Distillation Loss}

We explore two temperature settings~(1 and 3), and compare Jensen-Shannon~(JS) loss with the traditional Kullback-Leibler~(KL) in the 28k setup. Results are presented for validation~(Table~\ref{appendix:tab1}) and test sets~(Table~\ref{appendix:tab2}).

\begin{table}[h!]
\centering
\begin{tabular}{@{}lcc|ccc|ccc@{}}
\toprule
                & \textbf{FLEURS} & \textbf{CV-13} & \multicolumn{3}{c|}{\textbf{FLEURS}} & \multicolumn{3}{c}{\textbf{CV-13}} \\ \midrule
                & \textbf{avg}    & \textbf{avg}   & \textbf{ca}      & \textbf{ta}     & \textbf{th}     & \textbf{ca}    & \textbf{ta}    & \textbf{th}  \\ \midrule
JS w/ $\tau=1$ & \textbf{14.8}	& 8.8	& \textbf{12.7}	& 10.7 & 12.9	& \textbf{5.9}	&\textbf{18.7} & 9.9 \\
JS w/ $\tau=3$ & 15.4	& \textbf{8.5}	& 14.5	& \textbf{10.0}	& \textbf{12.7}	& 6.0	& 18.9	& \textbf{9.4} \\
KL w/ $\tau=1$ & 15.6	&10.2	&15.1	&13.8	&12.8	&6.3	&18.8	&10.5 \\
KL w/ $\tau=3$ & 15.7	&8.6	&15.2	&10.5	&13.0	&\textbf{5.9}	&18.8	&9.5      
\\ \bottomrule
\end{tabular}
\caption{WER~($\downarrow$) for in- and out-of-domain \textbf{validation sets} for \ourssp equipped with JS and KL losses, with different temperatures~($\tau$). Best results in \textbf{bold}.}
\label{appendix:tab1}
\end{table}

\begin{table}[h!]
\centering
\begin{tabular}{@{}lcc|ccc|ccc@{}}
\toprule
                & \textbf{FLEURS} & \textbf{CV-13} & \multicolumn{3}{c|}{\textbf{FLEURS}} & \multicolumn{3}{c}{\textbf{CV-13}} \\ \midrule
                & \textbf{avg}    & \textbf{avg}   & \textbf{ca}      & \textbf{ta}     & \textbf{th}     & \textbf{ca}    & \textbf{ta}    & \textbf{th}  \\ \midrule
JS w/ $\tau=1$ & \textbf{15.4}   & \textbf{9.3}   & \textbf{13.1}    & 19.2            & 14.0            & \textbf{11.3}  & \textbf{10.9}  & \textbf{5.7} \\
JS w/ $\tau=3$ & 16.3            & 9.7            & 14.8             & 20.1            & 14.1            & 11.8           & 11.3           & 5.9          \\
KL w/ $\tau=1$ & 15.6            & 10.8           & 14.6             & \textbf{18.7}   & \textbf{13.3}   & 14.9           & 11.3           & 6.2          \\
KL w/ $\tau=3$ & 16.5            & 9.7            & 15.8             & 19.8            & 14.0            & 12.2           & 11.1           & 5.9          \\ \bottomrule
\end{tabular}
\caption{WER~($\downarrow$) for in- and out-of-domain \textbf{test sets} for \ourssp equipped with JS and KL losses, with different temperatures~($\tau$). Best results in \textbf{bold}.}
\label{appendix:tab2}
\end{table}

We observe overall better validation scores using JS, a trend that is confirmed by our test results. Regarding temperature, JS equipped with~$\tau=3$ presents inferior validation scores, but superior CV-13~(in-domain) test scores.
In this work, we selected our models based on their performance on the validation set of CV-13, which is why we report results for JS with $\tau=1$ only.

% \section{Stability of the Method}
% 
% \input{tables/appendix_3}
% 
% We investigate the stability of the different proposed methods, CLSR-FT, \ourssp using JS with~$\tau=1$ and \ourssp using KL with $\tau=1$, by running three different seeds using the 28k setup. Results are presented in Table~\ref{appendix:tab3}. Considering a confidence interval of 95\%, we confirm \ourssp with JS loss is statistically superior than its version with only CLSR FT.\footnote{Confidence intervals are the following. \\CLSR-FT: [15.54, 15.87], JS:~[15.42, 15.49], KL:~[15.24, 15.57].}    
% 
% 
\section{Stability of the Methods}

\begin{table}[h!]
    \centering
    \resizebox{\textwidth}{!}{
    \begin{tabular}{ll|cccc|cccc|cccc}
    \toprule
    \multirow{2}{*}{Domain} & \multirow{2}{*}{Split}
    & \multicolumn{4}{c|}{\textbf{CLSR-FT}}
    & \multicolumn{4}{c|}{\textbf{JS} ($\tau=1$)}
    & \multicolumn{4}{c}{\textbf{KL} ($\tau=1$)} \\
    \cmidrule(lr){3-6}\cmidrule(lr){7-10}\cmidrule(l){11-14}
    && s1 & s2 & s3 & \textbf{avg}
    & s1 & s2 & s3 & \textbf{avg}
    & s1 & s2 & s3 & \textbf{avg} \\
    \midrule
    \multicolumn{14}{c}{\emph{Catalan (ca)}}\\
    \midrule
    \multirow{2}{*}{FLEURS} & Val  & 13.1 & 13.9 & 13.6 & \textbf{13.5}\scriptsize{$\pm$0.4}  & 12.7 & 12.6 & 13.1 & \textbf{12.8}\scriptsize{$\pm$0.3}  & 15.1 & 14.8 & 14.7 & \textbf{14.8}\scriptsize{$\pm$0.2} \\
    & Test & 13.5 & 13.3 & 13.6 & \textbf{13.5}\scriptsize{$\pm$0.2}  & 13.1 & 13.6 & 13.3 & \textbf{13.3}\scriptsize{$\pm$0.2}  & 14.6 & 14.6 & 14.3 & \textbf{14.5}\scriptsize{$\pm$0.2} \\
    \addlinespace
    \multirow{2}{*}{CV-13} & Val  &  9.6 &  9.6 &  9.0 & \textbf{ 9.4}\scriptsize{$\pm$0.4}  & 10.7 &  9.3 &  8.8 & \textbf{ 9.6}\scriptsize{$\pm$1.0}  & 13.8 & 13.8 & 13.7 & \textbf{13.8}\scriptsize{$\pm$0.1} \\
    & Test & 11.4 & 11.3 & 11.2 & \textbf{11.3}\scriptsize{$\pm$0.1}  & 11.3 & 11.3 & 11.3 & \textbf{11.3}\scriptsize{$\pm$0.03} & 14.9 & 14.6 & 15.0 & \textbf{14.8}\scriptsize{$\pm$0.2} \\
    \midrule
    \multicolumn{14}{c}{\emph{Tamil (ta)}}\\
    \midrule
    \multirow{2}{*}{FLEURS} & Val  & 19.2 & 18.9 & 19.0 & \textbf{19.1}\scriptsize{$\pm$0.2}  & 18.7 & 19.1 & 18.5 & \textbf{18.8}\scriptsize{$\pm$0.3}  & 18.1 & 17.7 & 17.7 & \textbf{17.8}\scriptsize{$\pm$0.2} \\
    & Test & 19.6 & 19.8 & 19.9 & \textbf{19.8}\scriptsize{$\pm$0.1}  & 19.2 & 19.2 & 19.6 & \textbf{19.3}\scriptsize{$\pm$0.2}  & 18.5 & 18.3 & 17.9 & \textbf{18.2}\scriptsize{$\pm$0.3} \\
    \addlinespace
    \multirow{2}{*}{CV-13} & Val  &  9.3 &  9.3 &  9.0 & \textbf{ 9.2}\scriptsize{$\pm$0.2}  &  9.9 &  9.3 &  8.9 & \textbf{ 9.4}\scriptsize{$\pm$0.5}  & 10.5 & 10.4 & 11.4 & \textbf{10.8}\scriptsize{$\pm$0.5} \\
    & Test & 11.3 & 11.4 & 11.2 & \textbf{11.3}\scriptsize{$\pm$0.1}  & 10.9 & 10.9 & 11.2 & \textbf{11.0}\scriptsize{$\pm$0.2}  & 11.4 & 11.2 &  9.8 & \textbf{10.8}\scriptsize{$\pm$0.9} \\
    \midrule
    \multicolumn{14}{c}{\emph{Thai (th)}}\\
    \midrule
    \multirow{2}{*}{FLEURS} & Val  & 13.8 & 13.6 & 13.5 & \textbf{13.6}\scriptsize{$\pm$0.2}  & 12.9 & 12.9 & 12.9 & \textbf{12.9}\scriptsize{$\pm$0.02} & 12.8 & 13.0 & 13.0 & \textbf{12.9}\scriptsize{$\pm$0.1} \\
    & Test & 13.8 & 13.8 & 14.2 & \textbf{13.9}\scriptsize{$\pm$0.2}  & 14.0 & 13.6 & 13.6 & \textbf{13.7}\scriptsize{$\pm$0.2}  & 13.3 & 13.6 & 13.6 & \textbf{13.5}\scriptsize{$\pm$0.1} \\
    \addlinespace
    \multirow{2}{*}{CV-13} & Val  &  6.4 &  6.0 &  5.7 & \textbf{ 6.0}\scriptsize{$\pm$0.3}  &  5.9 &  5.8 &  5.9 & \textbf{ 5.9}\scriptsize{$\pm$0.1}  &  6.3 &  6.5 &  6.3 & \textbf{ 6.4}\scriptsize{$\pm$0.1} \\
    & Test &  6.2 &  6.0 &  5.9 & \textbf{ 6.0}\scriptsize{$\pm$0.2}  &  5.7 &  5.8 &  5.8 & \textbf{ 5.7}\scriptsize{$\pm$0.1}  &  6.2 &  6.3 &  6.2 & \textbf{ 6.2}\scriptsize{$\pm$0.1} \\
    \midrule
    \multicolumn{14}{c}{\emph{Averages}}\\
    \midrule
    \multirow{2}{*}{FLEURS} & Val  & 15.4 & 15.5 & 15.4 & \textbf{15.4}\scriptsize{$\pm$0.04} & 14.8 & 14.9 & 14.8 & \textbf{14.8}\scriptsize{$\pm$0.1} & 15.3 & 15.1 & 15.1 & \textbf{15.2}\scriptsize{$\pm$0.1} \\
    & Test & 15.6 & 15.6 & 15.9 & \textbf{15.7}\scriptsize{$\pm$0.2}  & 15.4 & 15.5 & 15.5 & \textbf{15.5}\scriptsize{$\pm$0.03} & 15.5 & 15.5 & 15.2 & \textbf{15.4}\scriptsize{$\pm$0.2} \\
    \addlinespace
    \multirow{2}{*}{CV-13} & Val  &  8.4 &  8.3 &  7.9 & \textbf{ 8.2}\scriptsize{$\pm$0.3}  &  8.8 &  8.1 &  7.9 & \textbf{ 8.3}\scriptsize{$\pm$0.5}  & 10.2 & 10.2 & 10.5 & \textbf{10.3}\scriptsize{$\pm$0.2} \\
    & Test &  9.7 &  9.6 &  9.4 & \textbf{ 9.5}\scriptsize{$\pm$0.1}  &  9.3 &  9.4 &  9.4 & \textbf{ 9.3}\scriptsize{$\pm$0.1}  & 10.8 & 10.7 & 10.3 & \textbf{10.6}\scriptsize{$\pm$0.3} \\
    \bottomrule
    \end{tabular}}
    \caption{Val/Test WER~($\downarrow$) for in-domain~(CV-13) and out-of-domain~(FLEURS) across three methods: fine-tuning-only version (\texttt{CLSR-FT}), and distillation versions (\ours) with either Jensen-Shannon~(JS) loss, or traditional Kullback-Leibler~(KL) one, both with temperature~$\tau=1$. All experiments are repeated with three seeds (s1–s3) and we also report in \textbf{bold} the \textbf{mean}$\pm$std. Rows are organized by per-language scores and dataset averages.}
    \label{appendix:tab3}
\end{table}

We investigate robustness by repeating each approach — fine-tuned only (\texttt{CLSR-FT}), \ourssp with JS ($\tau=1$), and \ourssp with KL ($\tau=1$)—over three random seeds in the 28k setup. Table~\ref{appendix:tab3} reports per-language results and dataset averages as mean$\pm$std WER (↓) across seeds.

% Overall, variance across seeds is small. On out-of-domain FLEURS, JS shows particularly tight dispersion (e.g., Test average $15.5\pm0.03$), while KL and \texttt{CLSR-FT} remain within $\approx\!0.1\text{--}0.2$ absolute WER. In-domain CV-13 is similarly stable on Test (JS $9.3\pm0.1$, \texttt{CLSR-FT} $9.5\pm0.1$, KL $10.6\pm0.3$); the largest fluctuations appear on a few validation splits (e.g., Catalan CV-13 Val for JS, $9.6\pm1.0$, and Tamil CV-13 Test for KL, $10.8\pm0.9$), but these do not change the overall ranking.

With a 95\% confidence intervals on the macro-averaged scores across seeds, \ourssp with JS is statistically superior to the fine-tuning-only baseline, while KL is comparable to JS in mean but exhibits higher variability on some splits, with is probably related to the borderline effects reported in recent research \citep{js_vs_kd_acl23,js_vs_kd_bb2023}.\footnote{95\% confidence intervals over seed means: \texttt{CLSR-FT} [15.54,\,15.87], JS [15.42,\,15.49], KL [15.24,\,15.57].} In short, JS ($\tau{=}1$) offers the best trade-off between accuracy and stability across languages and domains, confirming that distillation improves over \texttt{CLSR-FT} with lower seed sensitivity.

\end{document}